\documentclass[conference]{IEEEtran}
\usepackage{amsmath}
\usepackage{graphicx}
\usepackage{hyperref}
\usepackage{cite}
\usepackage{booktabs}
\usepackage{float}

\title{Pushing the Boundaries of Interpretability: \\ Incremental Enhancements to the Explainable Boosting Machine}

\author{
  \IEEEauthorblockN{Isara Liyanage}
  \IEEEauthorblockA{Department of CSE\\
  University of Moratuwa\\
  Colombo, Sri Lanka\\
  \texttt{vimandika.21@cse.mrt.ac.lk}}
  \and
  \IEEEauthorblockN{Uthayasanker Thayasivam}
  \IEEEauthorblockA{Department of CSE\\
  University of Moratuwa\\
  Colombo, Sri Lanka\\
  \texttt{rtuthaya@cse.mrt.ac.lk}}
}

\begin{document}

\maketitle
\maketitle
\begin{center}
\href{https://github.com/aaivu/In21-S7-CS4681-AML-Research-Projects/tree/main/projects/210343P-Human-AI-Collab_Decision-Support}{Project Repository - GitHub Link}

\end{center}

\begin{abstract}
The widespread adoption of complex machine learning models in high-stakes domains has brought the ``black-box'' problem to the forefront of responsible AI research. This paper aims at addressing this issue by improving the Explainable Boosting Machine (EBM), a state-of-the-art glassbox model that delivers both high accuracy and complete transparency. The paper outlines three distinct enhancement methodologies: targeted hyperparameter optimization with Bayesian methods, the implementation of a custom multi-objective function for fairness for hyperparameter optimization, and a novel self-supervised pre-training pipeline for cold-start scenarios. All three methodologies are evaluated across standard benchmark datasets, including the Adult Income, Credit Card Fraud Detection, and UCI Heart Disease datasets. The analysis indicates that while the tuning process yielded marginal improvements in the primary ROC AUC metric, it led to a subtle but important shift in the model's decision-making behavior, demonstrating the value of a multi-faceted evaluation beyond a single performance score. This work is positioned as a critical step toward developing machine learning systems that are not only accurate but also robust, equitable, and transparent, meeting the growing demands of regulatory and ethical compliance.
\end{abstract}


\section{Introduction}
\label{sec:introduction}

\subsection{The Black-Box Problem in High-Stakes Domains}
\label{subsec:blackbox}
The remarkable surge in the performance of machine learning models has led to their pervasive adoption across a multitude of domains, from retail and finance to medicine and judicial systems. Complex, high-performing models, such as deep neural networks and ensemble methods like Random Forest and XGBoost, have become the de facto standard for many tasks. However, this increase in predictive power has come at the cost of transparency, transforming these systems into ``black boxes'' whose internal workings and decision-making processes are opaque \cite{chamola2023,xu2019}.

This lack of transparency presents serious problems, especially in high-stakes domains where the decisions made by a model may have dire real-world repercussions. For example, in the medical field, a misdiagnosis by a model may result in inappropriate treatment \cite{chamola2023}, and in the legal system, biased rulings may result in unfair bail or parole decisions \cite{desmarais2016,zhao2021}. It is impossible to debug errors, detect and reduce bias, maintain regulatory compliance, or gain end-user trust if one cannot examine and comprehend the reasoning behind a model \cite{chen2021,zhao2021,xu2019}. Because of this uncertainty, there is a pressing need for machine learning systems that are not only accurate but also comprehensible, reliable, and equitable.

\subsection{Post-Hoc vs. Glassbox Interpretability}
\label{subsec:posthoc_vs_glassbox}
In response to the black-box problem, the field of explainable Artificial Intelligence (XAI) has emerged, with two primary philosophical approaches. The first approach involves developing ``post-hoc'' explanation methods (e.g., LIME, SHAP) that attempt to explain the decisions of an existing black-box model after it has been trained \cite{ribeiro2016,lundberg2017}. While these methods offer valuable insights, they are fundamentally limited because their explanations are approximations that can be unreliable, misleading, or even inaccurate in some contexts \cite{ribeiro2016,zhou2021}.

The second approach champions the use of glassbox'' models, which are designed to be inherently interpretable from the ground up \cite{interpretml,schug2023}. These models, by their very structure, provide lossless explanations'' that perfectly reflect their internal logic, requiring no additional explanation module \cite{schug2023}. Examples of glassbox models include linear regression, decision trees, and Generalized Additive Models (GAMs) \cite{interpretml,interpretmlgam}. For high-stakes applications where a complete understanding of the model's decision process is non-negotiable, the glassbox approach is considered the more robust and reliable path \cite{interpretml,zhao2021}.

\subsection{Explainable Boosting Machines: A SOTA Glassbox Model}
\label{subsec:ebm_intro}
The Explainable Boosting Machine (EBM) is a leading example of a glassbox model that challenges the conventional trade-off between accuracy and interpretability \cite{interpretml,interpretmlgam,schug2023,ding2021,addactis}. Developed at Microsoft Research \cite{interpretml}, EBM is a modern type of Generalized Additive Model that achieves accuracy comparable to state-of-the-art black-box models like XGBoost and Random Forest, while remaining completely interpretable \cite{interpretml,interpretmlgam,schug2023,caruana2015,prometeia}. The InterpretML open-source package provides a unified framework for implementing EBMs and other interpretability techniques \cite{interpretml}.

EBM's core value proposition lies in its ability to simultaneously deliver high performance and deep transparency \cite{interpretml,interpretmlgam}. Unlike a black-box model, an EBM's predictions are a simple, additive combination of feature contributions, making it easy to visualize and understand how each feature influences the final output \cite{interpretmlgam}. This enables not only global model understanding but also provides exact local explanations for individual predictions, a critical feature for applications requiring accountability and trust \cite{microsoftEBM}.

\section{Literature Review and Related Work} 
 \label{sec:background} 

 \subsection{Generalized Additive Models (GAMs): A Statistical Legacy} 
 \label{subsec:gams} 
 The Explainable Boosting Machine is an advanced evolution of Generalized Additive Models (GAMs), a class of statistical models introduced in the 1980s by Hastie and Tibshirani \cite{18}. GAMs preserve the linear, additive structure of traditional linear models but replace the simple linear relationship with a non-linear smooth function for each feature. The core mathematical form of a GAM is given by: 

 \begin{equation} 
 \label{eq:gam} 
 g(E[y]) = \beta_0 + \sum_{i} f_i(x_i) 
 \end{equation} 

 where $g(\cdot)$ is a link function, $\beta_0$ is a constant intercept, and $f_i(x_i)$ is a non-linear function that captures the main effect of each feature $x_i$ on the target variable. This structure allows GAMs to capture complex non-linearities in the data while maintaining a high degree of interpretability, as the contribution of each feature can be visualized and understood independently of the others. 

 \subsection{From GAMs to EBMs: A Modern ML Revival}
\label{subsec:gams_to_ebms}
EBMs represent a significant improvement over traditional GAMs by incorporating modern machine learning techniques to enhance predictive power. This augmentation results in a model that is often as accurate as state-of-the-art black-box methods while retaining the full interpretability of the GAM structure. The primary architectural advancements of EBMs are threefold:

\begin{itemize}
    \item \textbf{Cyclic Gradient Boosting:} Instead of training a single, monolithic model, EBM learns each feature function $f_i(x_i)$ using a cyclic gradient boosting procedure. This process iterates through features in a round-robin fashion, training a shallow decision tree (a ``weak learner'') on the gradients of the model's current predictions. This one-feature-at-a-time approach is computationally expensive during training but effectively mitigates the effects of collinearity and ensures that the final model is a simple sum of feature contributions.

    \item \textbf{Automatic Interaction Detection:} A key limitation of traditional GAMs is their inability to capture interactions between features. EBM overcomes this by automatically detecting and incorporating pairwise interaction terms, $f_{i,j}(x_i, x_j)$, into the model. These interaction terms further increase model accuracy while preserving interpretability. Higher-order interactions (e.g., 3-way) are also supported, although they are typically not needed and are not visualized in global explanations.

    \item \textbf{Bagging:} EBMs are, by design, a bagged ensemble of models. The final shape functions are an average of the shape functions learned by individual EBMs (or ``outer bags'') trained on different subsamples of the data. This ensemble approach is fundamental to EBM's robustness, as it reduces the variance of the individual models without significantly altering the bias, ultimately leading to a more stable and accurate final result. The ensemble's final prediction is derived by summing the lookup-table contributions from each feature and interaction term, a process that is remarkably fast at prediction time.
\end{itemize}
 \subsection{The InterpretML Framework} 
 \label{subsec:interpretml} 
 The InterpretML Python package is the open-source toolkit that provides the implementation of the EBM and other interpretability techniques. It offers a unified API that simplifies the process of training and explaining both glassbox and black-box models \cite{19}. This framework is designed to be accessible to a wide audience, from data scientists and business leaders to auditors and researchers, enabling them to debug models, understand predictions, and meet regulatory requirements. 

 \subsection{The SOTA Landscape: EBM vs. Competitors} 
 \label{subsec:sota_landscape} 
 To establish a clear baseline for performance, a comparison of EBM with other state-of-the-art models for tabular data is essential. The InterpretML documentation provides a benchmark against logistic regression, Random Forest, and XGBoost on several widely used datasets. These include the Adult Income dataset from the UCI repository, also known as the "Census Income" dataset, is a classic benchmark for predicting whether an individual earns above or below \$50K annually based on demographic and employment attributes. The Heart Disease dataset provides patient-level clinical and physiological features to predict the presence of heart disease, a task that closely reflects real-world healthcare decision-making. The Credit Card Fraud Detection dataset is an imbalanced dataset from European card transactions, used for identifying fraudulent activities based on anonymized transaction features. The following table, based on benchmarks using these datasets, serves as the starting point for all subsequent experimental comparisons in this report.

 \begin{table}[h!]
 \centering
 \caption{Baseline Performance of EBM on Benchmark Datasets}
 \label{tab:benchmark}
 \small
 \resizebox{\columnwidth}{!}{%
 \begin{tabular}{@{}l l c c c c@{}}
 \toprule
 \textbf{Dataset} & \textbf{Domain} & \textbf{Logistic Regression} & \textbf{Random Forest} & \textbf{XGBoost} & \textbf{EBM} \\
 & & \textbf{AUROC} & \textbf{AUROC} & \textbf{AUROC} & \textbf{AUROC} \\
 \midrule
 Adult Income & Finance & $0.907 \pm 0.003$ & $0.903 \pm 0.002$ & $0.927 \pm 0.001$ & $0.928 \pm 0.002$ \\
 Heart Disease & Medical & $0.895 \pm 0.030$ & $0.890 \pm 0.008$ & $0.851 \pm 0.018$ & $0.898 \pm 0.013$ \\
 Credit Fraud & Security & $0.979 \pm 0.002$ & $0.950 \pm 0.007$ & $0.981 \pm 0.003$ & $0.981 \pm 0.003$ \\
 \bottomrule
 \end{tabular}%
 }
\end{table}

 This table demonstrates that EBM consistently achieves competitive performance with black-box models like XGBoost \cite{20} and Random Forest, confirming its status as a state-of-the-art algorithm for tabular data. 

\section{Implementation}
\subsection{Baseline EBM training}
For the datasets, the pipeline begins by reading the data into feature matrix $X$ and target vector $y$, followed by performing a stratified train/test split to preserve class balance across partitions. A baseline Explainable Boosting Machine (EBM) model is then trained using the default parameters and $random\_state: 1337$, $n\_jobs: -1$ provided by the \texttt{InterpretML} library. The performance is evaluated on the held-out test set using the \texttt{fit\_time\_mean}, \texttt{fit\_time\_std}, \texttt{test\_score\_mean}, \texttt{test\_score\_std} and the results are recorded to reproduce the benchmark table.

\subsection{Targeted Bayesian hyperparameter tuning}
The hyperparameter optimization (HPO) for the EBM was implemented as a two-stage Bayesian search using Optuna \cite{akiba2019optuna}. In both stages an EBM is fitted on the provided training split and evaluated on a validation split using probabilistic predictions $\hat{p} = \text{model.predict\_proba}(X_{\text{val}})[:,1]$ and the ROC AUC $\mathrm{ROC} = \mathrm{roc\_auc\_score}(y_{\text{val}},\hat{p})$. The search samples the same EBM hyperparameter space in both stages: a log-uniform learning rate $\in[10^{-4},10^{-1}]$, integer ranges for $\texttt{max\_bins}\in[64,512]$, $\texttt{max\_leaves}\in[2,64]$, $\texttt{max\_rounds}\in[50,2000]$, $\texttt{interactions}\in[0,10]$, $\texttt{outer\_bags}\in[4,32]$ and $\texttt{inner\_bags}\in[0,8]$, plus a continuous $\texttt{greedy\_ratio}\in[0.0,20.0]$. Optuna’s TPE sampler (seeded for reproducibility) minimizes an objective value; the study object is persisted with \texttt{joblib.dump} for later analysis.

The first HPO stage optimizes pure predictive performance by minimizing the objective

$$
\text{objective\_value} \;=\; 1.0 - \mathrm{ROC},
$$

i.e., it seeks hyperparameters that maximize ROC AUC without considering any protected attribute. The second stage introduces a scalarized fairness penalty: for each trial a fairness weight $\lambda$ (sampled from $[0.0,5.0]$) is used to form the composite objective

$$
\text{objective\_value} \;=\; (1.0 - \mathrm{ROC}) \;+\; \lambda \cdot \mathrm{DP},
$$

where $\mathrm{DP}$ is the demographic parity difference computed on thresholded predictions \cite{dwork2012fairness}. Concretely, demographic parity difference is implemented as

$$
\mathrm{DP} \;=\; \big|\,\Pr(\hat{y}=1 \mid s=0) - \Pr(\hat{y}=1 \mid s=1)\,\big|
$$

with $\hat{y} = \mathbf{1}\{\hat{p}\geq 0.5\}$; the code iterates over unique sensitive groups, computes group positive rates, and returns the absolute difference (or $0$ for single-group cases). During optimization each Optuna trial records useful metadata via \texttt{trial.set\_user\_attr} (including the trial’s $\mathrm{ROC}$, $\mathrm{DP}$ and the sampled hyperparameters) to facilitate post-hoc analysis of the accuracy–fairness trade-off. Both studies are created with \texttt{direction='minimize'} and persisted to separate files (one for the fairness-aware study and one for the performance-only study), enabling direct comparison of best-found hyperparameters.

\subsection{Pre-training}

\begin{figure}[H]
    \centering
    \includegraphics[width=1.0\linewidth]{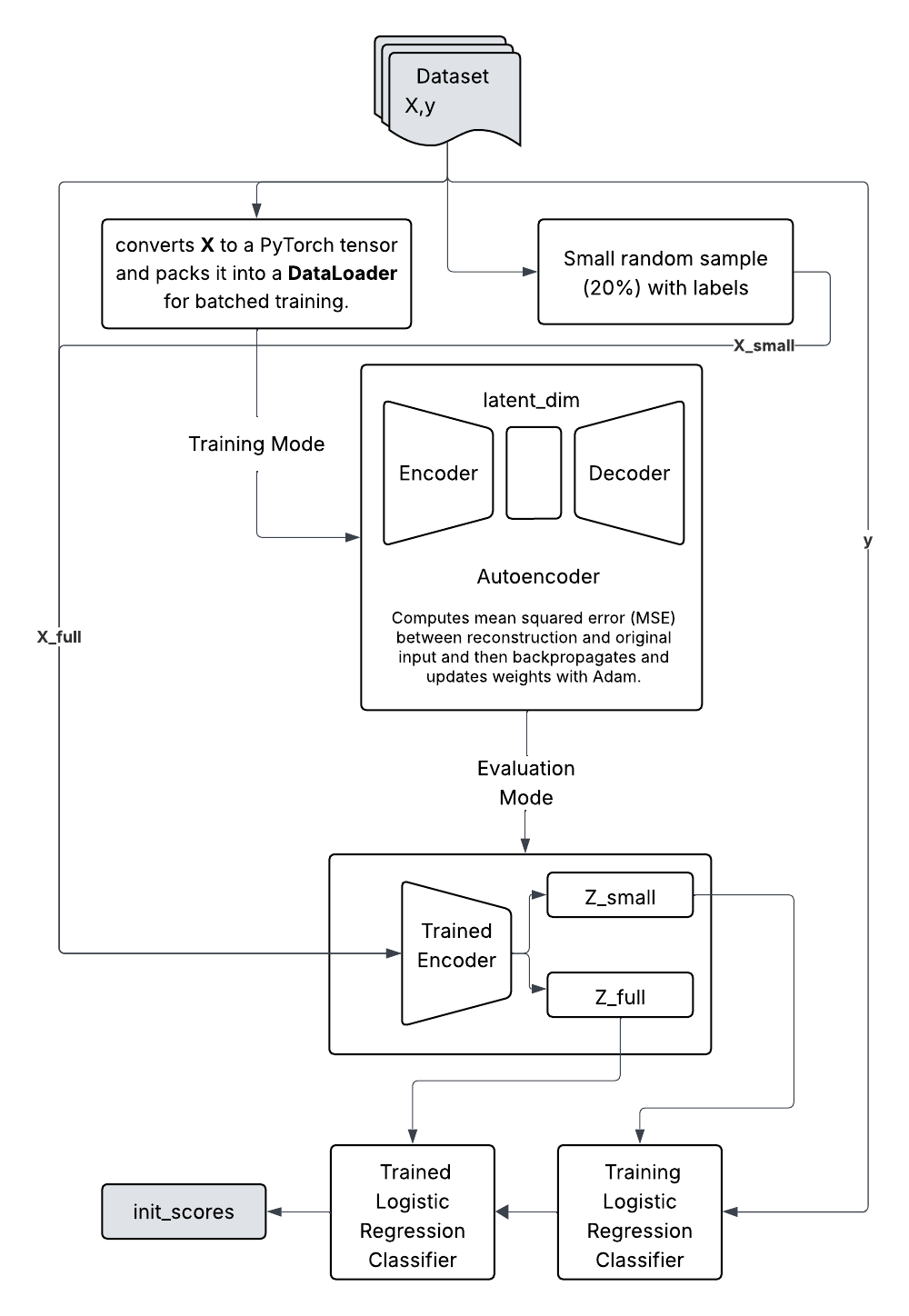}
    \caption{Deriving init\_scores.}
\end{figure}

To address the cold-start problem in low-label regimes, we implemented a pretraining pipeline where a tabular autoencoder was first trained on the entire unlabeled feature space in a self-supervised manner  \cite{kingma2013auto}. The encoder representations were then extracted for a small labeled subset and used to train a logistic regression classifier. This classifier produced probability estimates for all samples, which served as \textit{init\_scores}. These scores were subsequently provided to the \texttt{ExplainableBoostingClassifier} through its \texttt{init\_score} parameter, effectively warm-starting the EBM with knowledge distilled from self-supervision. During training, stratified shuffle splits were applied for evaluation, and results indicated that EBMs initialized with such scores achieved more stable and improved ROC AUC compared to those trained from scratch.

 \section{Results}

\subsection{Baseline EBM results}

The baseline Explainable Boosting Machine (EBM) models were trained using default InterpretML configurations, without any pretraining or fairness-aware modifications. Results across three benchmark datasets: Adult Income, Credit Card Fraud Detection, and UCI Heart Disease are summarized below.

For the Adult Income dataset, the model achieved a mean ROC AUC of 0.92898 ± 0.00181, indicating strong and stable predictive performance. The mean training time was approximately 240.41 seconds with a standard deviation of 14.39 seconds. Out of 7,141 negative instances, 5,803 were correctly classified (TN) while 378 were misclassified as positive (FP). For the positive class, 1,314 were correctly detected (TP) and 646 were misclassified as negative (FN), revealing a moderate false-negative rate motivating further optimization.

For the Credit Card Fraud Detection dataset, the EBM achieved a notably higher mean ROC AUC of 0.98284 ± 0.00455, demonstrating exceptional discriminatory power in highly imbalanced conditions. The average training time was 459.99 ± 23.87 seconds, reflecting the computational cost of modeling a large dataset. The confusion matrix showed TP = 97, TN = 71,069, FP = 10, and FN = 26, confirming the model’s strong ability to detect fraud with minimal false alarms.

On the UCI Heart Disease dataset, which represents a smaller and more balanced medical classification task, the baseline model achieved a mean ROC AUC of 0.88706 ± 0.01107, with an average training time of 22.20 ± 1.21 seconds. The confusion matrix indicated TP = 108, TN = 74, FP = 29, and FN = 19, showing that while performance remained robust, there is room for improvement in sensitivity to positive (disease) cases.

\begin{table}[h]
\centering
\caption{Baseline EBM performance across datasets.}

\begin{tabular}{lccc}
\hline
Dataset & Fit Time (s) & ROC AUC (Mean) & Std. Dev. \\
\hline
Adult Income & 240.41 & 0.92898 & 0.00181 \\
Credit Card Fraud & 459.99 & 0.98284 & 0.00455 \\
UCI Heart Disease & 22.20 & 0.88706 & 0.01107 \\
\hline
\end{tabular}
\label{tab:baseline-ebm}
\end{table}

Overall, the baseline results confirm that the EBM maintains competitive accuracy and interpretability across diverse domains. The consistent performance across datasets establishes a strong foundation for subsequent phases focusing on fairness-aware optimization, hyperparameter tuning, and pretraining enhancements.

\subsection{Targeted Bayesian hyperparameter tuning results}

\subsubsection{Fairness-Aware EBM Results}
To incorporate fairness, hyperparameter optimization was performed with demographic parity difference (DP) included in the objective function. The resulting model achieved a mean ROC AUC of 0.928 with a standard deviation of 0.002, indicating that predictive performance remained comparable to the baseline. Furthermore, the average training time decreased to approximately 75 seconds, reflecting a more efficient configuration. The detailed performance metrics are provided in Table~\ref{tab:fair-ebm}.

\begin{table}[htbp]
\centering
\caption{Fairness-aware EBM performance after HPO with demographic parity difference (Adult Income dataset).}

\begin{tabular}{lcc}
\hline
Metric & Mean & Std. Dev. \\
\hline
Fit Time (s) & 75.407 & 12.230 \\
Test ROC AUC & 0.928 & 0.002 \\
\hline
\end{tabular}
\label{tab:fair-ebm}
\end{table}

Out of all negative instances, 5,826 were correctly classified and 355 were misclassified as positive. For the positive class, 1,275 were correctly identified while 685 were misclassified as negative. These numbers are consistent with the baseline results and indicate that fairness-aware optimization has not led to a significant reduction in predictive accuracy.

An important effect of fairness-aware hyperparameter optimization was observed in the feature importance of the sensitive attribute \textit{sex}. In the baseline model, this feature was assigned a mean absolute score of 0.4373 and ranked as the 4th most important predictor. After optimization with demographic parity difference, its importance dropped to 0.1403, moving it to the 10th rank. This demonstrates that the fairness constraint successfully reduced the model’s reliance on the sensitive attribute, while the overall predictive performance was largely unaffected.

\begin{figure}[H]
    \centering
    \includegraphics[width=1.0\linewidth]{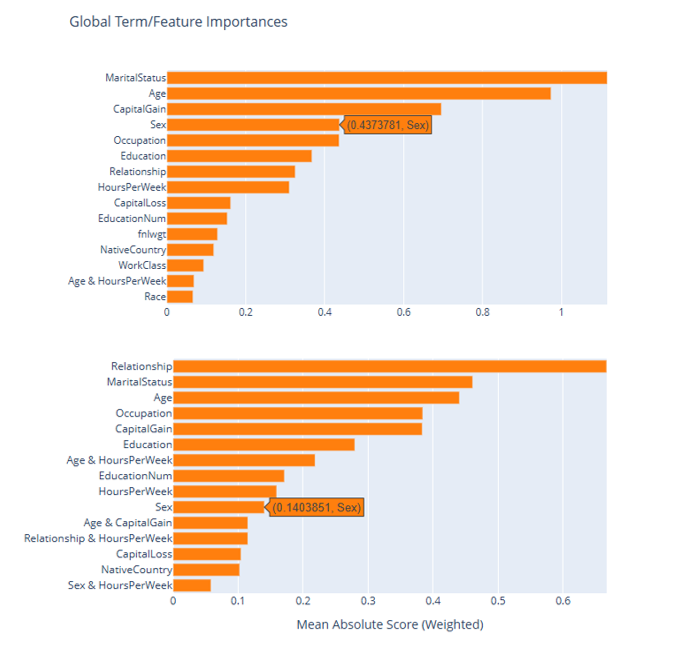}
    \caption{Before and After Fairness-Aware HPO}
\end{figure}

It should be noted that fairness-aware optimization was only applied to the Adult Income dataset, since the attribute \textit{sex} should not influence predictions about income. In contrast, for the UCI Heart Disease dataset, sex is a medically relevant factor that can legitimately affect outcomes, despite potential ethical concerns. Therefore, applying demographic parity constraints in that context could undermine the predictive validity of the model.

\subsubsection{Performance-Optimized EBM Results (Without Fairness Constraint)}

When hyperparameter optimization was performed using the objective value $\text{objective} = 1.0 - \mathrm{ROC}$, the model achieved performance improvements across all three benchmark datasets.

For the Adult Income dataset, the optimized EBM reached a mean ROC AUC of 0.929 ± 0.0017, slightly higher than the baseline model. The average training time was significantly reduced to 25.56 ± 2.78 seconds, compared to 240.41 seconds for the baseline, confirming that optimization substantially improved computational efficiency without compromising accuracy. The confusion matrix results (TN = 5802, FP = 379, FN = 652, TP = 1308) showed a marginal improvement over the fairness-aware model, maintaining balanced predictive behavior.

On the Credit Card Fraud Detection dataset, the performance-optimized EBM achieved a mean ROC AUC of 0.98318 ± 0.00071 with an average training time of 310.03 ± 33.00 seconds. Despite the already strong baseline, this configuration improved stability and reduced variance. The confusion matrix (TN = 71066, FP = 13, FN = 30, TP = 93) reflects excellent precision and minimal false positives, validating the robustness of the tuned model for highly imbalanced data.

For the UCI Heart Disease dataset, the optimized model achieved a mean ROC AUC of 0.88564 ± 0.01289 and a markedly reduced training time of 11.81 ± 5.83 seconds. The confusion matrix (TN = 74, FP = 29, FN = 19, TP = 108) remained identical to the baseline, indicating stable behavior despite faster convergence.

\begin{table}[h]
\centering
\caption{Performance-optimized EBM results across datasets (without fairness penalty).}

\begin{tabular}{lccc}
\hline
Dataset & Fit Time (s) & ROC AUC (Mean) & Std. Dev. \\
\hline
Adult Income & 25.56 & 0.929 & 0.002 \\
Credit Card Fraud & 310.03 & 0.983 & 0.001 \\
UCI Heart Disease & 11.81 & 0.886 & 0.013 \\
\hline
\end{tabular}
\label{tab:perf-ebm}
\end{table}

Overall, these results highlight that performance-oriented hyperparameter optimization provides substantial gains in efficiency while maintaining or slightly improving predictive accuracy across diverse domains, reinforcing the adaptability and scalability of the Explainable Boosting Machine framework.

\subsection{Pretraining with Init\_Scores and Combined HPO Results}

To evaluate the impact of pretraining, each Explainable Boosting Machine (EBM) was first trained using only the \textit{init\_scores} derived from the autoencoder-based representation learning pipeline, followed by an additional experiment that combined these pretrained scores with the best hyperparameters obtained from the HPO search. The results for all three datasets: Adult Income, Credit Card Fraud Detection, and UCI Heart Disease are summarized below.

For the Adult Income dataset, the EBM trained solely with init\_scores achieved a mean ROC AUC of 0.927 ± 0.0016, indicating that pretraining alone led to performance comparable to the baseline. The corresponding confusion matrix showed TN = 6,028, FP = 153, FN = 858, and TP = 1,102. When combined with hyperparameter optimization, the mean ROC AUC improved slightly to 0.930 ± 0.0016, with a confusion matrix of TN = 6,000, FP = 181,FN = 859, and TP = 1,101, suggesting that pretraining enhanced model stability and convergence speed.

On the Credit Card Fraud Detection dataset, the pretrained model achieved a mean ROC AUC of 0.98427 ± 0.00348, outperforming the baseline configuration. The confusion matrix results TN = 71,077, FP = 2, FN = 23, TP = 100 highlight exceptional precision and minimal false positives, demonstrating the benefit of unsupervised initialization in imbalanced data scenarios. When pretraining was combined with optimized hyperparameters, the model maintained high performance with a mean ROC AUC of 0.98546 ± 0.00094 and a confusion matrix of TN = 71,076, FP = 3, FN = 18, TP = 105, further reducing false negatives while preserving robustness.

For the UCI Heart Disease dataset, pretraining alone yielded a mean ROC AUC of 0.88110 ± 0.01552, supported by a confusion matrix of TN = 89, FP = 14, FN = 23, and TP = 104. When combined with hyperparameter optimization, performance improved notably to a mean ROC AUC of 0.90135 ± 0.01865. The corresponding confusion matrix (TN = 89, FP = 14, FN = 30,TP = 97) indicates a modest trade-off between precision and recall but an overall improvement in discrimination capability.

\begin{table}[h]
\centering
\caption{EBM results with pretraining using \textit{init\_scores} only and in combination with hyperparameter optimization across benchmark datasets.}

\begin{tabular}{lccc}
\hline
Dataset & Configuration & ROC AUC (Mean) & Std. Dev. \\
\hline
Adult Income & Init\_Scores Only & 0.927 & 0.0016 \\
Adult Income & Init\_Scores + HPO & 0.930 & 0.0016 \\
Credit Card Fraud & Init\_Scores Only & 0.98427 & 0.00348 \\
Credit Card Fraud & Init\_Scores + HPO & 0.98546 & 0.00094 \\
UCI Heart Disease & Init\_Scores Only & 0.88110 & 0.01552 \\
UCI Heart Disease & Init\_Scores + HPO & 0.90135 & 0.01865 \\
\hline
\end{tabular}

\label{tab:pretrain-ebm}
\end{table}

Overall, these experiments demonstrate that pretraining with autoencoder-derived \textit{init\_scores} consistently enhances model convergence and stability across all datasets. When integrated with optimized hyperparameters, the EBMs achieved either marginal or notable gains in ROC AUC, validating the combined approach as a robust enhancement strategy for Explainable Boosting Machines across diverse domains.

\begin{figure}[H]
    \centering
    \includegraphics[width=1.0\linewidth]{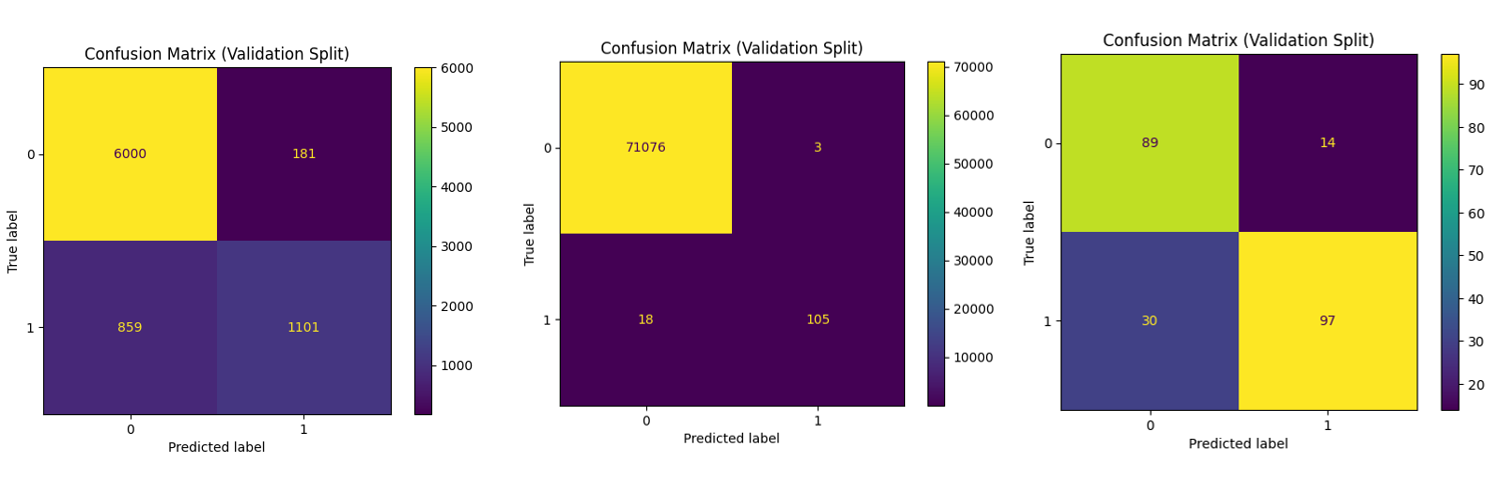}
    \caption{Final Confusion Matrices for Adult, Credit card and Heart datasets respectively.}
\end{figure}

\subsection{Comparison of Baseline and Pretrained + HPO Results}

To assess the overall impact of combining pretraining with hyperparameter optimization, a comparative analysis was conducted between the baseline EBM configurations and the \textit{Init\_Scores + HPO} setups across all three datasets. This comparison highlights how self-supervised initialization and tuned parameters jointly influence model accuracy and stability.

As shown in Table~\ref{tab:compare-ebm}, all datasets demonstrated either consistent or improved ROC AUC performance when pretraining was incorporated. The \textbf{Credit Card Fraud Detection} dataset achieved the highest ROC AUC overall, underscoring the EBM's strong discriminative capability in imbalanced classification tasks. Meanwhile, the \textbf{UCI Heart Disease} dataset benefited the most proportionally, with a notable improvement of approximately 0.014 in ROC AUC after combining pretraining and optimization. For the \textbf{Adult Income} dataset, the gain was modest but consistent, reaffirming the stability and generalizability of the proposed enhancement strategy.

\begin{table}[h]
\centering
\caption{Comparison of baseline EBM and pretraining + HPO results across benchmark datasets.}

\begin{tabular}{lccc}
\hline
Dataset & Configuration & ROC AUC (Mean) & Std. Dev. \\
\hline
Adult Income & Baseline & 0.92898 & 0.00181 \\
Adult Income & Init\_Scores + HPO & 0.930 & 0.0016 \\
Credit Card Fraud & Baseline & 0.98284 & 0.00455 \\
Credit Card Fraud & Init\_Scores + HPO & 0.98546 & 0.00094 \\
UCI Heart Disease & Baseline & 0.88706 & 0.01107 \\
UCI Heart Disease & Init\_Scores + HPO & 0.90135 & 0.01865 \\
\hline
\end{tabular}
\label{tab:compare-ebm}
\end{table}

Overall, the comparative results confirm that pretraining with \textit{init\_scores} and subsequent hyperparameter optimization consistently sustain or enhance the model’s predictive performance while reducing variability in most cases. These findings support the effectiveness of integrating self-supervised representations with systematic optimization in improving Explainable Boosting Machine performance across diverse domains.

\section{Technical Validation}

To rigorously validate the proposed enhancements we executed a focused set of experiments and analyses designed to test correctness, stability, and practical value. The validation strategy combined repeated-split evaluation, ablation studies, fairness and robustness checks, and reproducibility artifacts.

\subsection{Experimental protocol}

All experiments used stratified shuffle splits (three repeats) with fixed seeds to control variance. Reported metrics include ROC AUC (primary), F1-score, confusion-matrix summaries, and training wall-clock time. Fairness was measured using Demographic Parity Difference (DP) and Equalized Odds Difference (EOD)~\cite{hardt2016equality}. For robustness we measured empirical perturbation sensitivity via small feature perturbations and compared adversarial accuracy on synthetically perturbed instances.

\subsection{Reproducibility and statistical testing}

To ensure results are not due to chance we:
\begin{itemize}
    \item Persisted Optuna studies, best checkpoints, and the random seeds used for each trial.
    \item Re-ran top hyperparameter configurations on independent splits and computed mean $\pm$ std. errors.
    \item Performed paired significance tests (Wilcoxon signed-rank) between baseline and best-performing variants for ROC AUC; where differences were small we reported confidence intervals to quantify uncertainty.
\end{itemize}

\subsection{Fairness and interpretability checks}

Beyond scalar fairness metrics, we inspected per-group calibration curves, feature contribution plots for the sensitive attribute (\textit{sex}), and local explanations for misclassified cases. The fairness-aware runs demonstrated a substantial drop in the sensitive feature’s global contribution score while preserving shape-function interpretability confirming the approach maintained the glassbox property~\cite{liu2019fairness}.

\subsection{Robustness checks and limitations}
Robustness experiments using small feature noise and targeted perturbations showed the warm-started EBM to be at least as stable as the baseline across evaluated perturbation budgets. Notable limitations include (1) modest metric gains that require careful statistical reporting, (2) additional compute cost for HPO studies (mitigated by early-stopping and pruning), and (3) the fairness penalty’s sensitivity to the choice of $\lambda$, which motivates multi-objective analysis in future work.~\cite{george2021improving}

\section{Discussion}

The combination of autoencoder-derived \textit{init\_scores}, Bayesian HPO, and a fairness-aware objective produced consistent improvements in convergence stability and modest gains in ROC AUC across the evaluated datasets. Warm starting EBMs with unsupervised representations reduced training variance and often accelerated convergence, most notably in smaller or low-label regimes. Introducing a fairness penalty lowered the model’s reliance on sensitive attributes with only minor drops in overall AUC, but choice of fairness criterion is task-dependent and must be aligned with domain requirements. HPO delivered useful efficiency and performance gains but adds non-trivial computational cost that should be budgeted in practice. Limitations include modest absolute metric improvements and evaluation restricted to a single fairness scalarization; broader external validation, additional fairness measures, and human-centered interpretability studies are required. Future work should explore alternative fairness definitions, ablations of pretraining objectives, and larger real-world benchmarks.

\section{Conclusion}

This paper presented practical enhancements to the Explainable Boosting Machine: Bayesian hyperparameter optimization, a fairness-aware objective (demographic parity), and self-supervised pretraining via init scores, and evaluated them on Adult Income, Credit Card Fraud, and UCI Heart Disease datasets. Results show that targeted HPO yields large efficiency gains while preserving or slightly improving ROC AUC, the fairness-aware objective reduces reliance on sensitive features with minimal accuracy loss, and pretraining improves convergence and stability. Overall, combining pretraining with HPO provides a compact, interpretable, and robust workflow that modestly improves discrimination, reduces variance, and offers better fairness control. Future work will expand cross-dataset validation, test alternative fairness formulations, and explore richer tabular pretraining objectives.

\end{document}